\documentclass[conference]{IEEEtran}

\usepackage{hyperref}

\ifCLASSINFOpdf
   \usepackage[pdftex]{graphicx}
 
\else
 
\fi
\pdfoutput=1

\begin{document}
%
\title{Resolution of Unidentified Words in Machine Translation}

\author{\IEEEauthorblockN{Sana Ullah$^{\Psi}$, Md. Asdaque Hussain$^{\Psi}$, and Kyung Sup Kwak$^{\Psi}$}
\IEEEauthorblockA{$^{\Psi}$ Graduate School of IT and Telecommunications, Inha University\\
253 Yonghyun-Dong, Nam-Gu, Incheon 402-751, South Korea.\\
Email: sanajcs@hotmail.com, its\textunderscore asdaque@hotmail.com, kskwak@inha.ac.kr}}

%


\maketitle

\begin{abstract}
This paper presents a mechanism of resolving unidentified lexical units in Text-based Machine Translation (TBMT). In a Machine Translation (MT) system it is unlikely to have a complete lexicon and hence there is intense need of a new mechanism to handle the problem of unidentified words. These unknown words could be abbreviations, names, acronyms and newly introduced terms. We have proposed an algorithm for the resolution of the unidentified words. This algorithm takes discourse unit (primitive discourse) as a unit of analysis and provides real time updates to the lexicon. We have manually applied the algorithm to news paper fragments. Along with anaphora and cataphora resolution, many unknown words especially names and abbreviations were updated to the lexicon. 
\end{abstract}


%
\IEEEpeerreviewmaketitle

\section{Introduction}
A viable Machine Translation (MT) system must have a solution to anaphoric and elliptical ambiguities. It is not viable until it provides mechanism to handle the problem(s) of unidentified words such as names and abbreviations. Earlier, a discourse based approach is used to resolve anaphoric and elliptical ambiguities in Text-based Machine Translation (TBMT) \cite{1}. To produce high quality translation, the source text is dissected into mono-sentential discourses where complex discourses require further dissection either directly into primitive discourses or first into compound and later into primitive discourses \cite{2}. The resolution of ambiguities is performed during discourse processing stage \cite{3}. However, less attention has been given to resolve unknown words in TBMT. This paper improves dissection and discourse processing procedure by providing an algorithm for the resolution of unidentified lexical units. Firstly, we discuss discourse analysis and the dissection of complex and compound discourses into primitive discourses. Secondly, we present our proposed algorithm and analyze its validity by applying it to real world text, i.e., newspaper fragments. Finally, we present conclusion to our work.    

\begin{table*}[!t]
\renewcommand{\arraystretch}{1.3}
\caption{Primitive discourses and Generalized patterns (First complex discourse)}
\label{tab:1}
\centering
\begin{tabular}{|c|c|}
\hline
Primitive discourses &	Generalized patterns\\
\hline
IPEC of ILO &	A of B\\
\hline
IPEC has initiated &	A has B\\
\hline
Project is new &	A is B \\
\hline
(Time\textunderscore Bound\textunderscore Program) of (Government\textunderscore of\textunderscore Pakistan) is supported &	A of B is C\\
\hline
Elimination of WFCL	& A of B\\
\hline
Elimination from Pakistan &	A from C \\
\hline
\end{tabular}
\end{table*}

\begin{table*}[!t]
\renewcommand{\arraystretch}{1.3}
\caption{Primitive discourses and Generalized patterns (Second complex discourse)}
\label{tab:2}
\centering
\begin{tabular}{|c|c|}
\hline
Primitive discourses &	Generalized patterns\\
\hline
Project of IPEC &	A of B\\
\hline
IPEC will provide &	A will B\\
\hline
Assistance is technical &	A is B\\
\hline
Provided to the (Government\textunderscore of\textunderscore Pakistan) &	A to the B\\
\hline
(Convention\textunderscore 182) of ILO &	A of B\\
\hline
Implement on children &	A on B\\
\hline
Children are working &	A are B\\
\hline
Working conditions are hazardous &	A B are C\\
\hline
\end{tabular}
\end{table*}

\section{Discourse analysis and discourse unit}
The term discourse is introduced by Zellig Harris in 1952. A discourse is a connected piece of text of more than one sentence spoken by one or more speakers \cite{4}. Bellert defines discourse as a sequence of utterances $S_{1}, S_{2}, S_{3}....S_{n}$ such that the semantic interpretation of each utterance $S_{i}(2\leq i \leq n)$ is dependent on the interpretation of the utterances $S_{1}, S_{2}, S_{3}....S_{i-1}$ \cite{5}. However, interpretation of $S_{i}$ may depend on any subset of the set (say $U$) of previous utterances where $U=S_{1}, S_{2}, S_{3}....S_{n}$ \cite{2}. We think this definition needs further improvement in the context of cataphora resolution where interpretation of $S_{i}$ may sometimes depend on any subset of the set (say $\overline{U}$) of the subsequent utterances. A discourse unit is an atomic utterance that has no reference beyond its limitations or boundaries and can be mono-sentential or poly-sentential.     

\section{Dissection and Discourse processing}  
The entire dissection procedure is divided into two phases, i.e., dissection phase and discourse processing phase. In dissection phase, the source text is converted into primitive discourses. These primitive discourses are used to get generalized predicates. In discourse processing phase, various ambiguities are resolved including anaphoric and elliptical ambiguities. Mathematically, the dissection procedure can be represented as

\begin{equation}
T = (D_{1}, D_{2}, D_{3}......D_{i}...D_{l})l\geq1 
\end{equation}
\begin{equation}
T=\sum_{i=1}^l D_{i}
\end{equation}

where $T$= Source text and $D_{i}$= Poly or Mono sentential Discourse.

The resolution of the unidentified words is assumed to be processed during the discourse processing stage. The problem of unknown words resulted when we continued to apply the dissection concept to the real world text. For instance, consider the following newspaper fragment

\emph{[The International Labour Organization ILO's International Programme on the Elimination of Child Labour (IPEC) has initiated a new project to support the Government of Pakistan's Time Bound Programme on the elimination of the Worst Forms of Child Labour (WFCL) from Pakistan. The project will provide technical assistance to the Government of Pakistan to implement ILO Convention 182 on children working in hazardous working conditions.]}.

The two complex discourses are

1- [The International Labour Organization ILO's International Programme on the Elimination of Child Labour (IPEC) has initiated a new project to support the Government of Pakistan's Time Bound Programme on the elimination of the Worst Forms of Child Labour (WFCL) from Pakistan].

2- [The project will provide technical assistance to the Government of Pakistan to implement ILO Convention 182 on children working in hazardous working conditions].

The above discourses are dissected into the primitive discourses as given in Table \ref{tab:1} and Table \ref{tab:2}.

It is worth noting that one variable is used for the abbreviations as well as for the compound nouns. For example, Government of Pakistan is treated as a noun. Computer considers it noun by concatenating the text including some special characters (such as \textunderscore) defined by the programmer. Government of Pakistan is therefore considered as Government\textunderscore of\textunderscore Pakistan. Recognizing the correct noun is a challenging task for the computer/MT scientists. For instance, the Time Bound Program is considered noun from the previous discussion, but it might be treated otherwise. The decision must be based on the context. The concept of dissection is further explained in \cite{1}.   

\section{Unidentified lexical units} 

Creation of a list or glossary that contains well-known nouns, i.e., abbreviation and names, could be the best solution. The computer should add the new abbreviations or names (as they come along) to the list but sometimes the repetitive uses of nouns create problems for MT system. A pseudo code is written to resolve this problem. The pseudo code could be useful in Question Answering (QA) system, Information Retrieval (IR) system and in MT system, respectively.

\begin{figure*}[!t]
\centering
\includegraphics[width=5in]{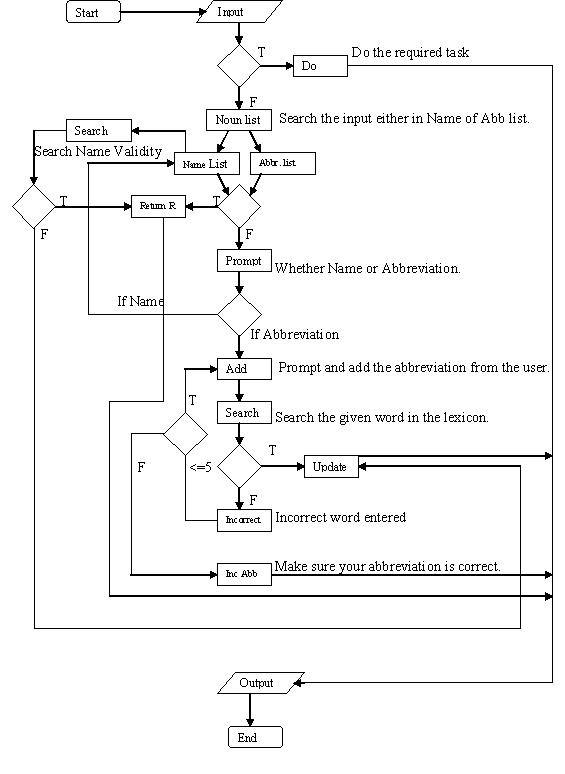}
\caption{Flow chart of the proposed algorithm}
\label{fig:1}
\end{figure*}

The proposed algorithm updates new names and abbreviations not present in the lexicon. Firstly, the existence of the noun is checked, i.e., whether the noun is correct or not. For instance, if a user enters Paksin Skidn Odind instead of Pakistan State Oil then the computer will subsequently reject it since these words not present in the lexicon. A problem could result if a user enters a word already in the lexicon, but not appropriate for the given abbreviation. For instance, the computer updates wrong abbreviation, if in case Pakistan Supreme Oil is entered instead of Pakistan State Oil. The best solution is that user is allowed to update inappropriate words for the given abbreviation, but these inappropriate words are less likely to be used by another user. Hence, the abbreviations that are less likely to be used are deleted automatically after a month. This could decrease the chances of errors. If these are not automatically deleted, then MT system could give invalid information, i.e., MT system may display Pakistan Supreme Oil instead of Pakistan State Oil.  The flow chart of the pseudo code is given in Fig \ref{fig:1}. It identifies the existence of noun in a lexicon. It takes primitive discourse as a unit of analysis. If noun doesn't exist, the nature of the noun either abbreviation or name is initially identified. If the noun is abbreviation/name, it is updated to the lexicon. Otherwise the user is requested to enter the required abbreviation/name for the updating purposes.

\section{Evaluation} 

The algorithm is applied to newspaper fragments and substantial numbers of unidentified lexical units are manually resolved. The evaluation of unknown words took place when we applied the dissection procedure to newspaper fragments . During our experiments, 124 names have been updated to the lexicon as compared to 57 abbreviations. Moreover, the number of unknown names appeared more than unknown abbreviations. However, updating new nouns depends on the lexicon. A poor lexicon results in considerable number of unidentified lexical units while a rich lexicon results in fewer unknown words.

\section{Conclusions} 

This paper explained the resolution of unidentified lexical units in TBMT. The resolution was considered as a part of discourse processing stage where apart from resolving other ambiguities, the resolution of unknown words was also considered. We presented an algorithm, which updates unknown nouns to the lexicon. The presented algorithm takes mono-sentential discourse as an input. The algorithm was manually applied to newspaper fragments and unidentified words were updated to the lexicon. In future this algorithm will be implemented as a part of our dissection model. Additionally, this could also be useful in QA and IR system.
\section*{Acknowledgement}

This research is supported by the MIC (Ministry of Information and Communication) South Korea, under ITRC (Information Technology Research Center) support program supervised by II TA  (Institute of Information Technology Advancement).
\end{document}